\pdfoutput=1

\documentclass[11pt]{article}

\usepackage[]{acl}

\usepackage{times}
\usepackage{latexsym}

\usepackage[T1]{fontenc}

\usepackage[utf8]{inputenc}
\usepackage{todonotes}

\usepackage{microtype}
\usepackage{amsmath,amssymb}
\usepackage{epsfig,graphicx,subfigure,caption}
\usepackage{algpseudocode}
\usepackage[normalem]{ulem}
\usepackage{amsmath}
\usepackage[linesnumbered,algoruled,boxed,noend]{algorithm2e}
\usepackage{xcolor}
\usepackage{color, colortbl}
\usepackage{multirow,booktabs, hhline}
\usepackage{amsmath, bm}
\usepackage[linesnumbered,algoruled,boxed,noend]{algorithm2e}
\usepackage{wrapfig}
\usepackage{listings}
\usepackage{adjustbox}

\usepackage{url}
\usepackage[T1]{fontenc}
\usepackage{tikz}
\usepackage{soul}
\usepackage{enumitem}

\setlist[enumerate]{itemsep=0pt, parsep=0pt}
\setlist[itemize]{itemsep=4pt, parsep=0pt}

\definecolor{darkgreen}{RGB}{0,100,0}
\definecolor{darkorange}{RGB}{184,85,0}
\definecolor{lightyellow}{RGB}{255,255,224}
\definecolor{loss}{RGB}{196,118,108}
\definecolor{win}{RGB}{164,193,136}

\newcommand{\ours}{SpeechFalcon\xspace}
\newcommand{\ourdata}{\textbf{\textsc{SpeechPref}}\xspace}
\newcommand{\lm}{ITLM}
\newcommand{\customnumbercircle}[3]{%
\tikz[baseline=(char.base)]{
\node[shape=circle, draw, inner sep=0.5pt, minimum size=3mm, fill=#1, text=#2, font=\scriptsize] (char) {#3};}}
\newcommand{\hlyellow}[1]{{\sethlcolor{lightyellow}\hl{#1}}}

\title{Speechworthy Instruction-tuned Language Models}

\author{
    Hyundong Cho\textsuperscript{1}\thanks{\xspace\xspace Work was done while HC was an intern at Amazon.},~
    Nicolaas Jedema\textsuperscript{2},~
    Leonardo F. R. Ribeiro\textsuperscript{2},~ 
    Karishma Sharma\textsuperscript{2},~ 
    \\
    \textbf{
    Pedro Szekely\textsuperscript{2},~ 
    Alessandro Moschitti\textsuperscript{2},~
    Ruben Janssen\textsuperscript{2},}~ and 
    \textbf{Jonathan May\textsuperscript{1}}
\\
\textsuperscript{1}University of Southern California, Information Sciences Institute~
\textsuperscript{2}Amazon
\\
{\small \texttt{hd.justincho@gmail.com}}
}

\begin{document}
\maketitle

\begin{abstract}

Current instruction-tuned language models are exclusively trained with textual preference data and thus are often not aligned with the unique requirements of other modalities, such as speech. 
To better align language models with the speech domain, we explore (\textit{i}) prompting strategies grounded in radio-industry best practices and (\textit{ii}) preference learning using a novel speech-based preference data of 20K samples, generated with a wide spectrum of prompts that induce varying dimensions of speech-suitability and labeled by annotators who \textbf{listen} to response pairs. 
Both human and automatic evaluation show that both prompting and preference learning increase the speech-suitability of popular instruction-tuned LLMs.
Interestingly, we find that prompting and preference learning can be additive; combining them achieves the best win rates in head-to-head comparison, resulting in responses that are preferred or tied to the base model in 76.2\% of comparisons on average. 
Lastly, we share lexical, syntactical, and qualitative analyses to showcase how each method contributes to improving the speech-suitability of generated responses.

\end{abstract}

\begin{figure}[ht!]
    \centering
    \includegraphics[width=\columnwidth]{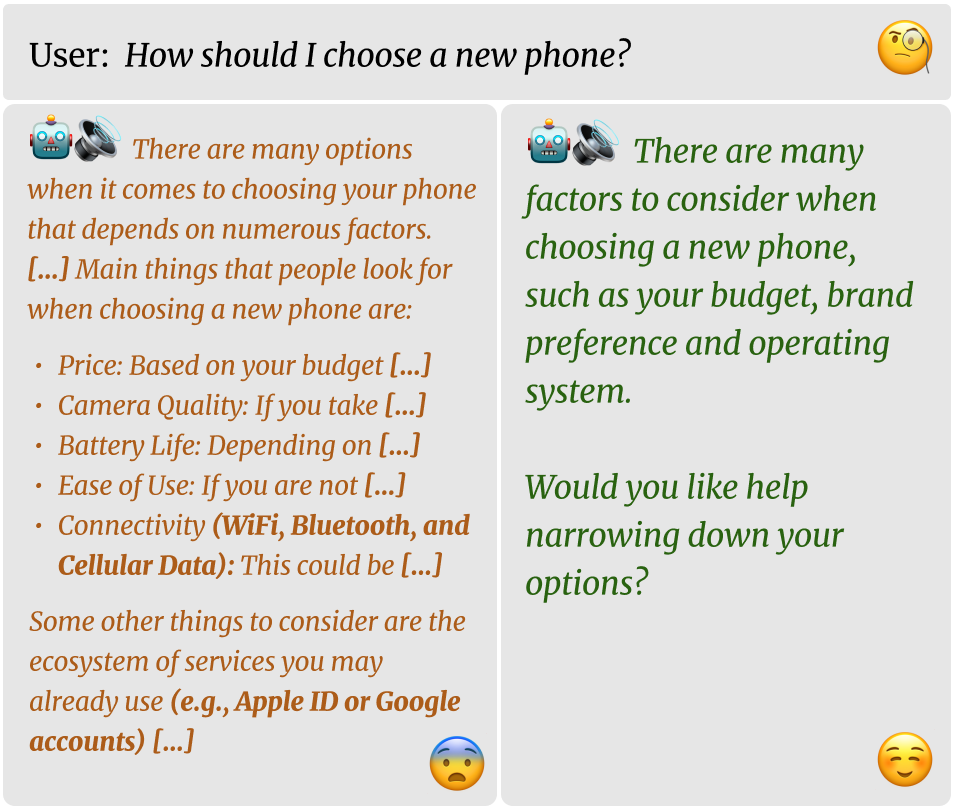}
    \caption{Current instruction-tuned language models tend to generate  \textcolor{darkorange}{\textbf{verbose}} responses with \textcolor{darkorange}{\textbf{non-vocalizable content}}, such as bullet lists or parentheses, that are not suitable for responses that are delivered as speech by voice assistants (\textcolor{darkorange}{left}, from OLMo 7B Instruct). Speech is serial and transient, and therefore \textcolor{darkgreen}{\textbf{concise yet informative}} responses with conversational \textcolor{darkgreen}{\textbf{follow-up questions}} are often preferred (\textcolor{darkgreen}{right}, from our adapted OLMo model).
    }

    \vspace{-1em}
    
    \label{fig:motivating_example}
\end{figure}

\section{Introduction}

Speech is one of our primary means of communication and a convenient and popular mode for interacting with virtual assistants~\cite{yang2004innovative}. 
Virtual assistants are a prime application of instruction-tuned language models (\lm), as both seek to provide helpful responses to user requests~\cite{peng2023instruction, chung2022scaling, wang2022self, wang-etal-2022-super, wei2021finetuned, sanh2022multitask, zhou2023lima}. 
However, current \lm s are fine-tuned on textual instructions ~\cite{peng2023instruction, chung2022scaling, wang2022self, wang-etal-2022-super, wei2021finetuned, sanh2022multitask, zhou2023lima} and preferences obtained from annotators that \emph{read} textual responses ~\cite{bai2022training, pmlr-v162-ethayarajh22a, ouyang2022training, touvron2023llama}.

We hypothesize that user preferences for speech and text are different and that the shift from generating written text to spoken language may pose a challenge for \lm s ~\cite{gonzalez-etal-2021-explanations}. 
Speech is serial and transient; speech processing is strictly linear  ~\cite{flowerdew1994academic} and requires higher cognitive load than reading ~\cite{thompson1996can, osada2004listening}. 
Concise and simple sentences, as seen on the right side of \autoref{fig:motivating_example}, are thus often preferred in speech~\cite{kern2008sound, abel2015out}, yet current \lm s optimized with textual preference datasets ~\cite{stiennon2020learning, singhallongway2023} produce verbose responses with non-vocalizable content (left side of \autoref{fig:motivating_example}). 

We test the hypothesis that current \lm s are sub-optimal for speech by surveying how user preferences differ when responses are heard instead of read. 
Our survey results reveal that users negatively perceive responses from current \lm s, primarily due to inadequate response length, information volume, and understanding ease.  
To improve \lm~speech-worthiness, we explore methods that improve \lm s responses in each of these dimensions. 

We explore two main methodologies to make these improvements: (i) prompt engineering~\cite{brown2020language, wei2021finetuned, kojima2022large} and (ii) preference learning~\cite{stiennon2020learning, bai2022training, rafailov2024direct}. 
For prompt engineering, we compile rules-of-thumb from the radio industry and findings from literature on listenability~\cite{messerklinger2006listenability, kotani-etal-2014-listenability} to design prompts for both zero-shot and few-shot in-context learning~\cite{kern2008sound, abel2015out, dowling2019immersive, kolodzy2012practicing, messerklinger2006listenability}. 
For preference learning, we collect the first speech-based preference dataset \ourdata, consisting of over 20K response pairs with varying degrees of speech-suitable responses.
To obtain responses that vary on the many dimensions of speech-suitability, we use multiple \lm s, generation hyperparameters, and varying system prompts that target different features (e.g. response length, simplicity, listen-ability). 
In addition, we designed a \emph{speech-only} preference annotation, in which annotators are only allowed to \textbf{listen} to the response pairs to accurately capture speech-based preferences. 
We use \ourdata to align models with the resulting speech-preference data using both proximal policy optimization (PPO)~\cite{schulman2017proximal} and direct preference optimization (DPO)~\cite{rafailov2024direct}. 

Our experiments 
with Falcon 7B Instruct~\cite{falcon40b} and OLMo 7B Instruct~\cite{groeneveld2024olmo}
show that both prompt-engineering and preference learning yields models that significantly improve over the base model on both human evaluations and automatic evaluations that serve as proxies to speech suitability. 
We show that these benefits are additive; combining both techniques produces responses that are maximally preferred to other setups, leading to an average of 75.3\% win or tie rate.
Further, we conduct lexical and syntactical analysis to show how our approaches adapt models for speech and share qualitative examples that compare responses from various techniques.
Lastly, we show that our prompts and in-context examples also benefit black box models like GPT-4, resulting in responses that are preferred or tied 88.3\% of the time compared to the unprompted model.
\footnote{We make our code, data, and trained models available at \url{https://justin-cho.com/speechllm}.}

\begin{table*}[t]
\small
\centering
\resizebox{\textwidth}{!}{
\renewcommand{\arraystretch}{1}
    \begin{tabular}{p{0.1\textwidth}p{0.15\textwidth}p{11cm}}
    \toprule
     & \textbf{Category} & \textbf{Prompt} \\ 
    \midrule 
    \multirow{3}{*}{\textbf{Kept}} & \textbf{Factual QA} & \textit{Why can camels survive for long without water?}  \\
    & \textbf{Brainstorming} & \textit{Give me some ideas to manage my manager}  \\ 
    & \textbf{Advice} & \textit{How do I start running?}  \\ 
    \midrule 
    \multirow{3}{*}{\textbf{Removed}} & \textbf{Creative Writing} & \textit{Write a scene between two actors discussing movie Inception.}  \\ 
    & \textbf{Summarization} & \textit{Summarize in one sentence this article about a famous song.} \{article\}  \\ 
     & \textbf{Info. Extraction} & \textit{Provide a list of all numbers mentioned and what they represent:} \{context\}  \\ 
     \bottomrule
    \end{tabular}}
    \caption{
        Examples of speech-suitable prompts that are kept and removed from Dolly-15K for evaluation and response sampling. Removed user prompts are those that are unlikely to be asked in a speech-based setting.
    }
    \label{tab:filtered_examples}
    \vspace{-1em}
\end{table*}

\begin{table}[t]
\centering
\resizebox{\columnwidth}{!}{
\renewcommand{\arraystretch}{1}
    \begin{tabular}{lrrr}
    \toprule
         \multirow{2}{*}{\textbf{Criteria}} & \multicolumn{3}{c}{\textbf{Score}} \\  \cmidrule(lr){2-4}
          & \multicolumn{1}{c}{\textbf{0}} & \multicolumn{1}{c}{\textbf{1}} & \multicolumn{1}{c}{\textbf{2}} \\
         \midrule 
         \textbf{Helpfulness} & Not helpful & Somewhat helpful & Helpful \\
         \textbf{Relevance} & Not relevant & Somewhat relevant & Relevant \\
         \textbf{Accuracy} & Not accurate & Contains minor errors & Accurate \\
         \textbf{Informative.} & Too little/much& Fair & Good \\ 
         \textbf{Length} & Too short/long & Short/Long & Adequate \\ 
         \textbf{Understand.} & Difficult & Fair & Easy \\ 
         \bottomrule
    \end{tabular}}
    \caption{We evaluate each response with various dimensions of speech-suitability using a 3 point Likert scale mapped from 0 to 2. Evaluation results with these dimensions for responses delivered in text and audio are shown in \autoref{tab:speech_vs_text_survey_results}.
    }

    \vspace{-1em}
    \label{tab:speech_vs_text_likert_scale}
\end{table}

\section{Speech-Suitability of \lm~Responses}
\label{sec:text_vs_audio_overall}

We initiate our study on the speech-suitability of \lm~ responses by answering these questions: (i) \textit{``Are \lm~ responses suitable for speech-based interactions?''} and (ii) \textit{``If not, what makes them unsuitable?''}

\subsection{Radio Industry Best Practices}
\label{sec:lessons_audio_reporting}

\citet{kern2008sound} and \citet{abel2015out} provide a glimpse to the answers for these questions by illustrating numerous examples of how news reporting and storytelling in text differs from audio. 
Through decades of radio experience, they establish rules-of-thumb to improve information delivery specific to the audio modality. 

We highlight rules that generalize beyond the news domain below, including: (\textit{i}) use simple words and sentence structures: allot a sentence to each idea and put the subject at the beginning as much as possible; (\textit{ii}) do not use atypical syntax, such as \textit{``President Bush today told members of congress''} and \textit{``I today went shopping''}; (\textit{iii}) avoid hyphenated adjectives (e.g. \textit{mineral-rich}, \textit{tech-heavy}); (\textit{iv}) avoid too many names and numbers; and other minor ones such as (\textit{v}) avoid tongue twisters and (\textit{vi}) avoid too much alliteration. 
Similar principles have been echoed in podcasting~\cite{dowling2019immersive}, multimedia journalism~\cite{kolodzy2012practicing}, and literature on listenability~\cite{chall1948predicting, fang1966easy, messerklinger2006listenability}.

\subsection{Text vs. Speech for \lm s}
\label{sec:text_vs_audio}

Speech guidelines from applications like radio and podcasting may or may not generalize to virtual assistant application, as the latter is more interactive and involve a nonhuman interlocutor.  
Therefore, we verify whether best practices from radio also apply for speech-based interactions with \lm s by comparing how listeners perceive response suitability in different modalities. 

First, we filter the user-prompt and response pairs from the Dolly-15K dataset~\cite{DatabricksBlog2023DollyV2} by removing prompts that require additional external context other than the request itself or those that explicitly ask for code or long form outputs that are unsuitable requests for speech-based interactions. 
We show categories and examples that were kept and removed in \autoref{tab:filtered_examples}. 
To conduct our survey, we randomly sampled 80 prompts from this subset, generated text responses with GPT-4 (\texttt{gpt-4-0613}), and converted them to speech using the generative engine models from Amazon Polly\footnote{\url{https://docs.aws.amazon.com/polly/latest/dg/generative-voices.html}}.
Each spoken response was evaluated by 3 annotators, resulting in a total of 240 annotations.
Since we are measuring speech-based preferences, the annotators can only \textbf{listen} to the response for the audio task. 
We report other details with regards to these annotations, including inter-annotator agreement and our annotation interfaces in Appendix \ref{sec:annotation_interface}. 

Following current practice in evaluating a \lm's response, we ask annotators to indicate the helpfulness, accuracy, and relevance for the response~\cite{ouyang2022training, zhao-etal-2022-fine-grained}. 
We additionally asked how suitable the responses were in terms of verbosity (length), information volume (informativeness), and comprehension ease (understandability). 
Each of these criteria was measured on a three-point Likert scale with scores mapped to the range 0 to 2, as shown in \autoref{tab:speech_vs_text_likert_scale}.
Note that informativeness and length have multiple options that correspond to the same score, such as `too little' or `too much', because we treat them as equally undesirable. 

\begin{table}[t]
\begin{adjustbox}{width=\columnwidth,center}

    \centering
    \begin{tabular}{lrrrrrr}
    \toprule
         &  \textbf{Rel.} & \textbf{Help.} & \textbf{Correct.} & \textbf{Inform.} & \textbf{Len.} & \textbf{Under.} \\  \midrule
         \textbf{Text} & 1.98 & 1.95 & 1.97 & 1.72 & 1.70 & 1.84 \\  
         \textbf{Audio} & 2.00 & 1.95 & 1.99 & \textbf{1.59}$^{\dagger}$ & \textbf{1.44}$^{\ddagger}$ & \textbf{1.77}$^{\dagger}$   \\ \bottomrule
    \end{tabular}
\end{adjustbox}
    \caption{Comparison between text and audio responses indicate the need to reduce excess information, decrease length, and enhance how easy the response is to comprehend when spoken. $^{\dagger}$ and $^{\ddagger}$ indicates a statistically significant drop compared to text results with a pairwise t-test at $p<0.05$ and $p<0.01$, respectively.}
    \label{tab:speech_vs_text_survey_results}

    \vspace{-1em}

\end{table}

\begin{table*}[t]
\begin{adjustbox}{width=\textwidth,center}

    \centering
    \begin{tabular}{p{\textwidth}}
    \toprule
         \footnotesize{\textbf{System Prompt:}\texttt{
         You are a helpful, respectful and honest voice assistant. 
         Respond \customnumbercircle{darkgreen}{white}{1} \hlyellow{colloquially using simple vocabulary and sentence structures.}
         \customnumbercircle{darkgreen}{white}{2} \hlyellow{Avoid jargon, hyphenated adjectives, excessive alliteration, and tongue twisters.} 
         Keep your response \customnumbercircle{darkgreen}{white}{3} \hlyellow{compact without missing key information}. 
         If a complete answer requires multiple steps, 
         \customnumbercircle{darkgreen}{white}{4} \hlyellow{provide only the first 3-5 and ask if the user is ready to move on} to the next steps or know more. 
         \customnumbercircle{darkgreen}{white}{5} \hlyellow{Make sure the response does not contain parentheses, numbered or bullet lists, and anything else that cannot be verbalized.} 
         The response should be \customnumbercircle{darkgreen}{white}{6} \hlyellow{suitable for speech such that it can be easily verbalized} by a text-to-speech system and understood easily by users when heard.  
         }}  \\ 
    \bottomrule
    \end{tabular}
\end{adjustbox}

    \caption{
    Our detailed system prompt is grounded in radio industry best practices outlined in \textsection \ref{sec:lessons_audio_reporting} (\customnumbercircle{darkgreen}{white}{1}, \customnumbercircle{darkgreen}{white}{2}), the findings from our survey in \textsection \ref{sec:text_vs_audio} (\customnumbercircle{darkgreen}{white}{3}, \customnumbercircle{darkgreen}{white}{4}), and refinements from manual iterations to remove other non-vocalizable content (\customnumbercircle{darkgreen}{white}{5}). Lastly, it provides the general purpose of the generated text to elicit an \lm's understanding of speech-suitability to cover any other unmentioned edge cases (\customnumbercircle{darkgreen}{white}{6}).     
    }
    \label{tab:detailed_prompt}

    \vspace{-1em}
\end{table*}

The annotation results in \autoref{tab:speech_vs_text_survey_results}, 
reveal that preferences for relevance, helpfulness, and correctness do not significantly differ by modality. 
However, preferences for information volume and length vary significantly between speech and text.
Decoupling the mappings that led to lower scores such as ``too little/short'' and ``too much/long'' in \autoref{tab:speech_vs_text_likert_scale} reveals that excess information and verbosity were the main culprits for lower scores (87.5\% and 90\%, respectively).
We additionally observe a significant decrease in the understandability of spoken responses, a result in line with the higher cognitive load required for speech processing~\cite{thompson1996can, osada2004listening}.
In conclusion, we demonstrate that some radio-industry best practices do generalize to the application of \lm s for speech and that user preferences for the speech and text modalities do indeed differ.
Further we reveal that the primary opportunity to improve LLMs for speech is to generate more concise responses that contain minimally sufficient information and are easier to understand.

\section{Adapting \lm s for Speech}
\label{sec:adaptation_methodology}

Having established the shortcomings of \lm~ responses, now we explore how to adapt \lm s to generate more speech-suitable responses.
In this section, we describe our approaches with two main directions: prompt engineering and preference learning.

\subsection{Prompt Engineering}
\label{sec:prompt_engineering}

Arguably the simplest method to generate speech-suitable responses is to specify an instruction or system prompt that induces the \lm's generation process towards the desiderata of speech ~\cite{raffel2020exploring, wang-etal-2022-super}.
This is effective because most recent language models are further fine-tuned to follow instructions~\cite{wei2021finetuned, sanh2021multitask} and is appealing because it has zero or very low data requirements. 
Guided by the rules-of-thumb from the radio industry (\textsection \ref{sec:lessons_audio_reporting}) and our findings from \textsection \ref{sec:text_vs_audio}, we iteratively refined system prompts using 20 randomly sampled input prompts (\textsection \ref{sec:text_vs_audio}) until LLMs generated more consistently speech-worthy responses.
We share our final detailed system prompt in \autoref{tab:detailed_prompt}, annotated with mappings to desiderata listed from \textsection \ref{sec:text_vs_audio_overall}.

A simple extension to a system prompt is to include examples that enable in-context learning (ICL), which have shown to further improve performance across many tasks~\cite{dong2023survey, gupta-etal-2023-coverage}. 
While our system prompt is descriptive, it can benefit a model to see actual examples of speech-suitable responses. 
To get a diverse set of in-context examples, we sample a few user prompts from each prompt category in Dolly-15K and generate responses using our detailed system prompt. 
Then, we edit these to use simpler, more colloquial language, as well as removing any non-vocalizable content if necessary. 
The set of in-context examples we used are shared in \autoref{sec:in-context_examples}. 

\begin{figure*}
    \centering
    \includegraphics[width=\textwidth]{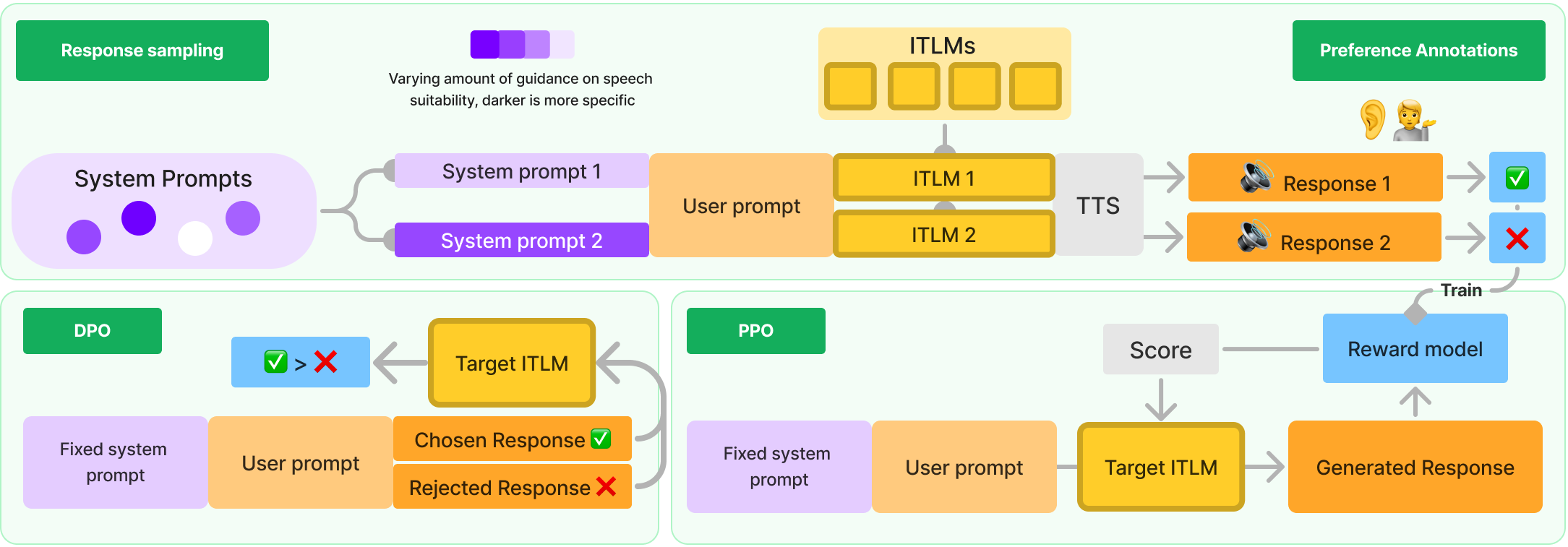}
    \caption{Preference learning method overview. Since we only have an approximate idea of what makes a good spoken response, we first compile a set of system prompts intended to vary the speech suitability of generated responses. We sample a pair to generate responses from various \lm s to further diversify responses, transform them to speech with a TTS service, and human annotators rank their preferences after listening. 
    } 
    \label{fig:method_overview}
    \vspace{-1em}

\end{figure*}

\subsection{Preference Learning}
\label{sec:rl_with_speech}

With prompt engineering, we are prescribing what accounts for speech-suitability.
Although these prompts are grounded in our findings from \textsection \ref{sec:text_vs_audio}, it is unlikely to an exhaustive guideline that generalizes to all types of prompts and audiences. 
Therefore, we investigate whether we can take a data-driven approach with user preferences to learn to generate speech-worthy responses. 
To this end, we collect a speech-based preference dataset and use preference learning algorithms to fine-tune models. 

\subsubsection{Response Sampling}
\label{sec:response_sampling}
Creating a reward model that robustly approximates speech-suitability requires collecting responses of varying quality across the different dimensions of speech-suitability.
To diversify response quality, we empirically compile a set of system prompts that can diversify generated responses. 
We pair these system prompts with various \lm~ s and generation hyperparameters to synthetically collect responses that follow the insights from \textsection \ref{sec:text_vs_audio} at varying levels of detail.
We cycle through each pair of configurations to generate responses that will be compared to one another using the filtered user prompts from Dolly-15K described in \textsection \ref{sec:text_vs_audio}.
We repeat this process until we collected 20K response pairs.
We share further technical details of our sampling process in Appendix \ref{sec:response_sampling_details}.

\subsubsection{Annotating Speech-based Preferences}
\label{sec:preference_annotations}

Similar to the annotation setup for \textsection \ref{sec:text_vs_audio}, we only allow the annotators to listen to the user prompt and responses in order to accurately capture speech-based preferences. 
They must listen to both completely before they're able to indicate preferences.
To minimize any bias introduced by listening order, the compared responses are presented in random order.
The preference annotation interface and the instructions for the annotations are shown in Appendix \ref{sec:annotation_interface}. 

As done for Llama-2~\cite{touvron2023llama}, annotators are asked to choose one response over another with the following choices: \texttt{significantly better}, \texttt{better}, \texttt{slightly better}, and \texttt{negligibly better}, which can be chosen to indicate ties. 
In addition, we ask for a brief explanation for their choice for quality control and qualitative analysis purposes. 
We name the 20K preference pairs that we collected with this procedure as \ourdata.
Further implementation details on annotator recruitment and inter-annotator agreement are shared in Appendix \ref{sec:annotator_details}.

\subsubsection{Fine-tuning}
\label{sec:reward_modeling_rlhf}

To fine-tune models with \ourdata, we leverage two popular preference learning methods: PPO~\cite{schulman2017proximal} and DPO~\cite{rafailov2024direct}. 

For PPO, we train a reward model that generates a scalar score for speech-suitability given a single user prompt and response pair. 
Following the procedure of \citeauthor{ouyang2022training}, we add a projection layer to the same model architecture and model weights that we plan to adapt with reinforcement learning and train with the pairwise binary ranking loss: 
$$\mathcal{L}_{ranking} = -\text{log}(\sigma(r_\theta(x,y_c))-r_\theta(x,y_r)),$$
where $r_\theta(x,y)$ is the reward model's score for the user prompt $x$ and the generated response $y$, given model weight $\theta$. $y_c$ is the response chosen by the annotator and $y_r$ is the rejected one. 
We use this trained reward model for PPO. 

DPO is an RL free method that learns from preference data by learning to maximize the difference between the language model's predicted probabilities of the chosen texts and those of the rejected texts. 
It is a much simpler method than PPO, whose training stability is highly sensitive to hyperparameters  and checkpoint selection strategies. 
An overview of these methods are shown in \autoref{fig:method_overview}.

Lastly, since there is no theoretical constraints that prevent us from appending the prompts that we developed earlier in \textsection \ref{sec:prompt_engineering}, 
we can easily combine prompting and preference learning by appending the prompts while fine-tuning with preferences. 
We only need to set the fixed system prompt shown in the bottom half of \autoref{fig:method_overview} with the detailed prompt or the ICL prompt. 
Therefore, we also examine this combined approach. 

\section{Experimental Setup}

\subsection{Models}
\label{sec:model}

We conduct our experiments with Falcon 7B Instruct~\cite{falcon40b} and OLMo 7B Instruct~\cite{groeneveld2024olmo} as our base models; henceforth, we refer to these as Falcon and OLMo respectively. 
We chose these models because they were the best-performing \lm s with an Apache 2.0 license at the time of our study.  

To simplify notations going forth, 
we add a suffix \textbf{\textsc{-Prompt}} for models using the detailed prompt in \autoref{tab:detailed_prompt} and \textbf{\textsc{-ICL}} for those that also include the in-context learning examples. 
Models trained with \ourdata using PPO/DPO add the suffix \textbf{\textsc{PPO}} and \textbf{\textsc{DPO}}, along with \textbf{\textsc{-Prompt}} and \textbf{\textsc{-ICL}} if both prompting and preference learning were used. 
For example, Falcon DPO-ICL indicates a Falcon model trained with DPO while using the ICL prompt. 
All our models are trained using LLaMA-Factory~\cite{zheng2024llamafactory}. 
For PPO, we train separate reward models for each Falcon and OLMo, which enhances memory efficiency during reinforcement learning through parameter sharing of the value model and the policy model. 
We share other technical details of our implementations and hyperparameters for training in Appendix \ref{sec:implementation_details}.

\subsection{Data}
\label{sec:data}

The user prompts that we use for sampling responses for preference annotations and the PPO step are from the filtered version of Dolly-15K~\cite{DatabricksBlog2023DollyV2}, described in \textsection \ref{sec:text_vs_audio}.
This instruction dataset does not have a predefined train-test split, and therefore we take a 9:1 train-test split. 
It is one of the first open source, human-written instruction dataset with a permissive creative commons license. 
\ourdata is used to train the reward model for PPO and for DPO training.

\subsection{Evaluation}
\label{sec:evaluation}

Our primary goal is to generate responses that are more frequently and significantly preferred than those from the base model. 
The \textit{base model} refers to the setup of generating response from the non-finetuned model  with our simplest prompt \textit{``You are a helpful, respectful, and honest voice assistant.''} 
We also compare our best-performing approaches with one another to discover which configuration leads to the best results. 
Another desired outcome is that responses from one of our approaches will be preferred over for the original responses (denoted as \textit{Original response}) as the latter were collected through a textual interface that does not consider speech suitability.

\paragraph{Human evaluation.} 
Similar to the human evaluation in \citet{zhou2023lima}, we perform a head-to-head comparison of responses from the target models using the same setup as the preference annotations for \ourdata (\textsection \ref{sec:preference_annotations}). 
We count \texttt{negligibly better} as ties while all others lead to a win or loss. 
For both OLMo and Falcon, \textbf{\textsc{-Prompt}}, \textbf{\textsc{-ICL}}, \textbf{\textsc{DPO}}, and \textbf{\textsc{-DPO-ICL}} variants are compared to the base model to measure how much stronger each is against the base model. 
To further solidify the relative performance, we compare the best performing models between the prompt engineering approach and the combined approach. 
Each comparisons involve 60 response pairs generated with a fixed random subset of user prompts from the test set described in \textsection \ref{sec:response_sampling}. 
We ensure that these have not been seen during preference learning or selected as in-context examples.

\paragraph{Automatic Evaluation.} With automatic evaluation, we analyze the relationship between speech preferences and reasonable measurements of criteria relevant to speech-suitability, as outlined in \textsection \ref{sec:text_vs_audio_overall}: listenability, sentence complexity, length, and vocalizable content. 
For listenability, we measure the Flesch Reading Ease (FRE) score, which estimates readability as a function of the number of syllables within each sentence; FRE is known to be highly correlated with listenability~\cite{chall1948predicting, fang1966easy, messerklinger2006listenability, kotani-etal-2014-listenability}. 
To quantify sentence complexity, we use SpaCy's dependency parser\footnote{\url{https://spacy.io/api/dependencyparser}} and measure the depth of the resulting dependency graph. 
For length, we simply split sentences on white space and count the total number of words. 
For vocalizable content, we measure the average number of generated nonvocalizable (NV) characters, such as \texttt{{([-])}|*\_\&\%\$\#@`\/} and non-ascii characters, included in the generated response. 
Lastly, we score responses with a trained reward model (RM). 
To ensure that this RM is not biased for PPO models that were optimized for them, we train a reward model that has not been used for any PPO with a GPT-J 6B model~\cite{gpt-j}.

\section{Results and Discussion}

\begin{figure*}[t]
    \centering
    \includegraphics[width=\textwidth]{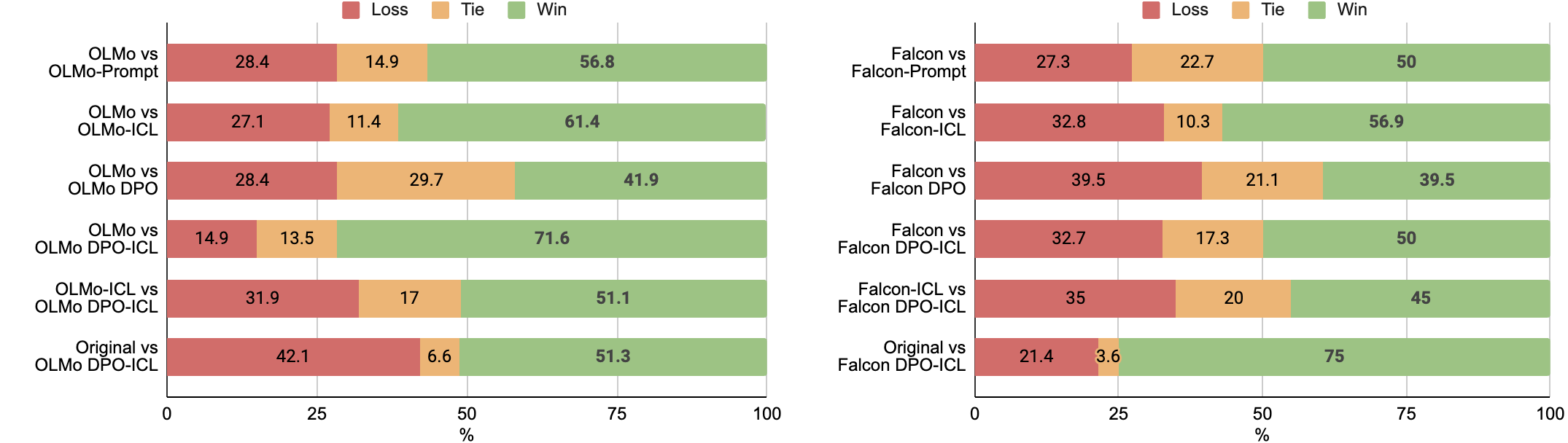}
    \caption{Head-to-head human evaluation results for OLMo (left) and Falcon (right). If the \textbf{\textcolor{win}{win}} rate is higher than the \textbf{\textcolor{loss}{loss}} rate, the model mentioned second in the y-axis ($B$ for $A$ vs $B$) is more often preferred in a speech setting. 
   }
   \vspace{-0.5em}
    \label{fig:human_eval_main}
\end{figure*}

\subsection{Human evaluation}
\label{sec:human_evaluation_results}

\paragraph{All techniques achieve significant win rates and the combined technique performs the best.} 

Our human evaluation results with OLMo and Falcon are shown in \autoref{fig:human_eval_main}, respectively. 
The first four rows in both figures clearly show that our techniques achieve a significantly higher win rate compared to the base model, with at minimum 13.5\% higher win rate and a maximum of 56.7\%. 
While most results are similar, OLMo DPO-ICL gets a higher win rate over the base model than OLMo-ICL does, but the opposite is true for Falcon. 
To evaluate their relative performance, we conduct a head-to-head comparison between DPO-ICL and ICL. 
With this setup, we find that DPO-ICL is more preferred than ICL for both OLMo and Falcon by 19.2,\% and 10\% respectively. 

DPO-ICL outperforming ICL suggests that the gains from prompt engineering and preference learning are additive. 
We suspect that DPO-ICL is able to improve over ICL because the ICL prompt helps the fine-tuning process by guiding the model towards speech-suitable responses from the very start of training and lets the preference learning data better refine the model's understanding of what is meant by speech-suitability. 
This dynamic is supported by DPO's training trajectories, an example of which is shown in \autoref{fig:dpo_training_trajectory}. 
It shows the prompted variants reaching a lower evaluation loss faster, achieving a higher evaluation accuracy sooner, and achieving larger margins between the probabilities of the chosen and rejected responses. 

We observe similar results with PPO overall and a head-to-head comparison between PPO vs DPO setups are close, with DPO narrowly edging PPO for OLMo and the two tying for Falcon. 
Therefore, for preference learning methods, we focus on DPO results and share our discussions on the comparison between PPO and DPO and details of PPO results in \autoref{sec:ppo_vs_dpo}.

\subsection{Automatic evaluation}
\label{sec:automatic_eval_results}

\begin{table}[t]
    \begin{minipage}{1\columnwidth}
	\centering
	\small
 	\resizebox{1\columnwidth}{!}{
		\begin{tabular}{lccccc}
            \toprule
            \multirow{2}{*}{Model} & \multirow{2}{*}{Word count $\downarrow$} & \multicolumn{2}{c}{Comprehensibility} & \multirow{2}{*}{RM $\uparrow$} & \multirow{2}{*}{NV $\downarrow$}  \\ \cmidrule(lr){3-4}
             & & FRE $\uparrow$ & DD $\downarrow$ \\
            \midrule 

            Original &  $69.07_{12.09}$ & $58.64_{2.88}$ & $6.43_{0.29}$ & $0.49_{0.14}$ & $1.8_{0.46}$ \\
            \midrule 

            OLMo & $211.10_{15.51}$ & $51.05_{2.49}$ & $6.64_{0.21}$ & $-0.38_{0.13}$ & $9.19_{1.37}$ \\
            \xspace\xspace -Prompt & $141.42_{14.66}$ & $\mathbf{53.56}_{2.16}$ & $6.61_{0.23}$ & $0.03_{0.15}$ & $5.92_{0.89}$ \\
            \xspace\xspace -ICL & $109.49_{10.20}$ & $48.97_{2.54}$ & $7.22_{0.21}$ & $0.31_{0.17}$ & $3.27_{0.57}$ \\
            \xspace\xspace  DPO & $174.82_{16.21}$ & $50.85_{2.08}$ & $\mathbf{6.56}_{0.18}$ & $-0.09_{0.15}$ & $5.69_{0.79}$ \\

            \xspace\xspace  DPO-ICL & $\mathbf{95.88}_{9.01}$ & $51.58_{2.57}$ & $7.23_{0.26}$ & $\mathbf{0.58}_{0.20}$ & $\mathbf{2.9}_{0.69}$ \\

            \midrule

            Falcon & $44.71_{10.26}$ & $49.74_{10.07}$ & $6.71_{0.51}$ & $0.41_{0.18}$ & $0.5_{0.18}$ \\
            \xspace\xspace -Prompt & $\mathbf{40.22}_{3.83}$ & $68.89_{2.46}$ & $\mathbf{6.06}_{0.22}$ & $0.76_{0.18}$ & $0.57_{0.18}$ \\
            \xspace\xspace -ICL & $47.76_{3.89}$ & $61.50_{2.36}$ & $6.10_{0.25}$ & $0.86_{0.16}$ & $0.42_{0.12}$ \\
            \xspace\xspace DPO & $41.49_{2.13}$ & $65.37_{2.73}$ & $6.38_{0.23}$ & $0.87_{0.20}$ & $\mathbf{0.34}_{0.10}$ \\ 
            \xspace\xspace DPO-ICL & $50.82_{3.21}$ & $\mathbf{69.52}_{1.81}$ & $6.52_{0.17}$ & $\mathbf{1.42}_{0.19}$ & $0.82_{0.26}$ \\

            \midrule
            \midrule

            GPT-4 & $182.41_{12.28}$ & $52.54_{1.45}$ & $6.22_{0.12}$ & $0.01_{0.12}$ & $35.06_{3.97}$ \\
            \xspace\xspace -Prompt & $\mathbf{37.07}_{1.43}$ & $\mathbf{75.1}_{1.54}$ & $\mathbf{5.73}_{0.12}$ & $1.85_{0.13}$ & $0.45_{0.1}$ \\
            \xspace\xspace -ICL & $39.57_{1.33}$ & $71.02_{1.44}$ & $6.11_{0.11}$ & $\mathbf{1.98}_{0.12}$ & $\mathbf{0.34}_{0.07}$ \\
            \bottomrule
        \end{tabular}
       }
    \end{minipage}
	\caption{Automatic evaluation results shed  light into factors that contribute to responses preferred for speech. FRE: Flesch Reading Ease, DD: Dependency Depth, RM: score from \ourdata-trained reward model, NV: nonvocal character count. $\downarrow$ means lower the better in general, while $\uparrow$ means higher the better in general. Best results for each dimension are in bold.
    }
	\label{tab:automatic-table}
        \vspace{-0.3cm}
\end{table}

\paragraph{Speech-worthiness is not simply about generating shorter and more readable responses.}
Automatic evaluation results are shown in \autoref{tab:automatic-table}, and we find that most results align with our expectations in that the more preferred responses are either shorter or easier to understand, and contains fewer nonvocalizable content. 
This trend is more pronounced for OLMo,  which originally generates longer responses compared to Falcon, and thus have higher chances of containing nonvocalizable content. 
On the other hand, Falcon already generates short responses without any adaptations, and interestingly its best performing variant, Falcon DPO-ICL, actually generates longer responses but with significantly improved FRE and RM scores. 

Syntactical complexity as measured by DD are also divergent in that it is lowered for Falcon but increased for OLMo in most cases. 
RM scores are the most consistent for both models and correspond well with results from human evaluations, but they suggest that DPO-variants are much more superior than the prompt-only methods when in fact their margins are not that large in human evaluations. 
As such, whether one model would be more preferred than another cannot be fully captured by these metrics, but they provide interesting insights into how models change based on our explored techniques. 

\subsection{Experiments with GPT-4}

We also experiment with GPT-4 (\texttt{gpt-4-turbo}) to examine whether our prompts can improve speech-suitability for black box models. 
Also, we compare our best results against GPT-4-Prompt/ICL to understand the shortcomings of smaller sized models and provide qualitative analyses in \autoref{sec:qualitative_analysis}. 

\begin{figure}[t]
    \centering
    \includegraphics[width=\columnwidth]{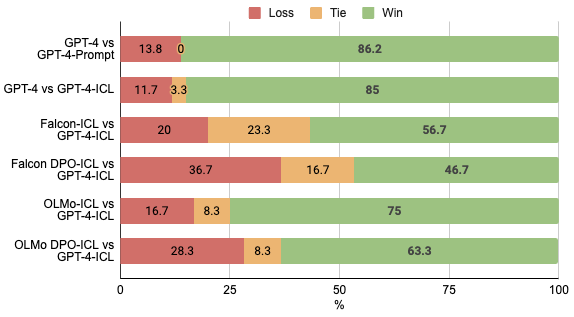}
    \caption{Head-to-head human evaluation results with our prompts using GPT-4. This figure takes the same format as \autoref{fig:human_eval_main}.
   }

   \vspace{-1em}
    \label{fig:gpt_results}
\end{figure}

\paragraph{GPT-4 also improves significantly with prompting and outperforms Falcon and OLMo models.}
The first two rows of \autoref{fig:gpt_results} show that the both \textbf{\textsc{-Prompt}} and \textbf{\textsc{-ICL}} significantly outperform GPT-4, and the automatic evaluation results shown in \autoref{tab:automatic-table} suggests that most of these gains are attributable to the almost 80\% reduction in length and 50\% increase in readability. 
We also see that the amount of nonvocalizable content is reduced to less than 1\% of what it used to generate. 
These improvements are further reflected in the reward score, which is the highest of all models we experimented with. 
The highest reward score is further reflected in the comparisons with our best performing Falcon and OLMo models. Falcon DPO-ICL is the model that comes closest, but it still falls short in its win rate by 10\%.

\section{Related work}

\subsection{Language models and speech}

Language models have been widely adopted for modular components of a voice assistant, mainly for automatic speech recognition~\cite{yu2016automatic, wang2020fairseq, chiu2018state}, response generation~\cite{cho-may-2020-grounding, zhou-etal-2022-reflect, liu-etal-2023-recap}, response selection~\cite{humeau2019poly, gao-etal-2020-dialogue, cho2021viola}, and speech synthesis~\cite{tan2022naturalspeech, wang2017tacotron, le2023voicebox}. 
While speech synthesis focuses on how to translate text to speech such that it \textit{sounds} natural, our work explores how to best compose the response itself for speech-based interactions. 
\citet{gonzalez-etal-2021-explanations} established that shorter explanations were preferred for open-domain question answering, but we find more nuanced preferences for \lm~responses.
More similar in context to our work, \citet{mousavi2024llms} explores whether \lm~ s are robust to speech-based interactions by analyzing their robustness to input that include automatic speech recognition (ASR) errors. 

Another line of research concerns developing multi-modal \lm s that can process both speech and text input~\citep{huang2023audiogpt}. 
\citet{zhang-etal-2023-speechgpt} trains a GPT-based model called SpeechGPT with both speech data and text data such that it does not requiring additional ASR or TTS systems.
However, this line of work also overlooks on how we should compose responses that are delivered as speech.

\subsection{Preference learning for \lm s}

While the paradigm of pre-training and then fine-tuning has become the defacto status quo, 
there is still active research in how to go about fine-tuning to get the best results and do it efficiently. 
One of the central methods for fine-tuning \lm s is through preference learning, which is a process of fine-tuning models with human preference data~\cite{bai2022training, pmlr-v162-ethayarajh22a, ouyang2022training, touvron2023llama}. 
PPO~\cite{schulman2017proximal} was one of the first to be successfully applied to natural language processing in a summarization task~\cite{stiennon2020learning}, but it is difficult to apply because of the challenge of attaining a reward model that is robust to distribution shifts. 
DPO~\cite{rafailov2024direct} has been proposed as an theoretically justified RL-free alternativee and has become popular due to its simplicity. 
Since then, many variants of DPO have appeared, such as SimPO~\cite{meng2024simpo}, IRPO~\cite{pang2024iterative}, KTO~\cite{ethayarajh2024kto}, but examining their effectiveness for \ourdata and developing a new preference learning method is not the scope of this work.

\section{Conclusion}

We explore an important yet overlooked challenge of adapting \lm s to compose responses that are specifically designed to be verbalized, i.e., speech-suitable. 
With rules-of-thumb of the radio industry and through our surveys that compare the suitability of a response for both text and audio, we establish that adaptations for speech-suitable responses is necessary for the broader application of \lm s to speech settings. 
To initiate this line of research, we explore prompting and preference learning with a unique speech-based preference we collected. 
Both human and automatic evaluations show that both methods are useful, and combining them yields to additive benefits.
Further, we showed that speech-suitability is a nuanced factor that is not fully explained by shorter and more easily understandable responses. 
Our work has focused on single-turn interactions and the content of the responses, and we hope to expand this investigation for multi-turn interactions and the acoustic aspect of speech-suitability.

\section*{Limitations} 

In this work, we focused on \textit{what} \lm s should generate for responses that will be delivered via speech. 
However, we recognize that another interesting line of research is \textit{how} the response should be verbalized, where factors related to speech, such as timber, pitch, and speed, are important. 
In addition, our examination is focused on single-turn interactions, but another intriguing realistic dimension to suitability of a response in speech is multi-turn interactions. 
It would be interesting to compare the effect of delivering information at various granularities and how different types of follow-up questions on the user experience with a \lm s in speech-based interactions. 
We leave these lines of research to future work. 

In addition, we were limited to the set of models that we investigated due to legal constraints that prevented us from experimenting with open-source models without an Apache 2.0 license. 
We look forward to the community without these constraints to the next steps in this line of research for further improving speech suitability in \lm s.

\section*{Broader Impact}

Since those who cannot read due to illiteracy or blindness rely on voice assistants to interface with modern technology, expanding their capabilities can directly lead to improvements in their standard of living. 
However, current voice assistants are not as generally useful as current state-of-the-art \lm s. 
The likes of Siri and Alexa tend to fulfill simple routine tasks and are brittle when facing complex requests. 
Therefore, adopting \lm s as the main backbones of voice assistants and adapting them to become suitable to interface through voice and can significantly increase the accessibility for these people to the powerful capabilities of \lm s.

\bibliography{anthology,custom}
\bibliographystyle{acl_natbib}

\appendix

\section*{Appendix}
\label{sec:appendix}

\begin{footnotesize}
\begin{table*}[t]
\begin{adjustbox}{width=\textwidth,center}

    \centering
        \begin{tabular}{m{0.3\textwidth} >{\footnotesize\ttfamily}m{0.9\textwidth}}
        \toprule
        \textbf{Prompt} & \multicolumn{1}{c}{\textbf{Response}} \\ \midrule
        How old is the Greek language? & The Greek language is over 3,500 years old and one of the oldest documented languages in the Western world. Would you like to learn more about the history of the Greek language? \\ \midrule
        Give me 3 different things that a smartphone can do better than a landline phone. & Smartphones can do many things that landline phones can't. They can access the internet, send text messages, and run apps. Would you like to learn more about other differences between smartphones and landline phones? \\ \midrule
        What to ask when interviewing doulas? & When interviewing doulas, you might want to ask about their training and certifications, how many births they've attended as a doula, and their philosophy and approach to supporting birthing families. You should also ask about how they handle unexpected situations, their availability around your due date, and their fee structure and the services they provide. You may also ask for references or testimonials from previous clients. Remember, it's important to find a doula who aligns with your values and makes you feel comfortable and supported during this important time. \\ \midrule
        Is it safe to invest in cryptocurrency? & It depends on the specific cryptocurrency you are considering investing in. Cryptocurrencies are volatile, and there is significant risk involved. It's important to do your own research and carefully consider your options before making any investment decisions. I can't provide investment advice but I can share more information about specific cryptocurrencies if you would like.\\ \midrule
        What are five capital cities in Asia? & Some of the capital cities in Asia include Tokyo, Japan; Beijing, China; New Delhi, India; Seoul, South Korea; and Bangkok, Thailand. Would you like to learn more about these cities or other capital cities in Asia? \\
        \bottomrule
        \end{tabular}
\end{adjustbox}
    \caption{
        In-context examples that we use for our prompting methods.  
    }
    \label{tab:in-context-examples}
\end{table*}
\end{footnotesize}

\begin{footnotesize}
\begin{table*}[t]
\begin{adjustbox}{width=\textwidth,center}

    \centering
        \begin{tabular}{m{0.15\textwidth} >{\footnotesize\ttfamily}m{0.9\textwidth}}
        \toprule
        \textbf{Prompt Name} & \multicolumn{1}{c}{\textbf{Prompt}} \\ \midrule
        Simple & You are a helpful, respectful and honest voice assistant. \\ \cmidrule(lr){2-2}
        Medium & You are a helpful, respectful and honest voice assistant. Make your responses colloquial and compact. Avoid any artifacts that cannot be vocalized. \\ \cmidrule(lr){2-2}
        Easy & You are a helpful, respectful, and honest voice assistant. Your response should use simple vocabulary and sentence structures so that it is easy to understand and such that it is below a 5th grade reading level. \\ \cmidrule(lr){2-2}
        Detailed & You are a helpful, respectful and honest voice assistant. Keep your response compact without missing key information and make it suitable for speech such that it can be easily verbalized by a text-to-speech system. Your response should use simple vocabulary and sentence structures. Avoid jargon, hyphenated adjectives, excessive alliteration, and tongue twisters. It should not contain parentheses, numbered or bullet lists, and anything else that cannot be vocalized. If a complete answer requires multiple steps, provide only the first 3-5 and ask if the user is ready to move on to the next steps or know more. \\
        \bottomrule
        \end{tabular}
\end{adjustbox}
    \caption{
        The set of system prompts that we use in Section \ref{sec:response_sampling} for sampling a variety of responses using Falcon Instruct 7B, GPT-3.5, and GPT-4. 
    }
    \label{tab:system_prompts}
\end{table*}
\end{footnotesize}

\section{In-context examples}
\label{sec:in-context_examples}

The in-context examples that we used in our experiments are shown in \autoref{tab:in-context-examples}.

\begin{figure*}
    \centering
    \includegraphics[width=\textwidth]{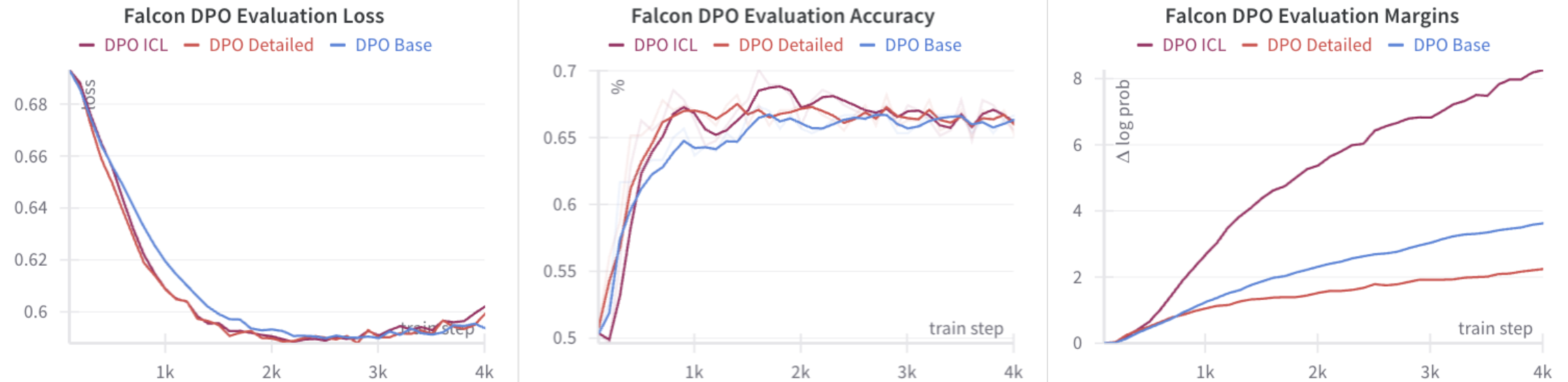}
    \caption{Falcon's DPO training trajectory suggests that prompts help the preference learning process by providing useful initial guidance for distinguishing between chosen and rejected responses.  
}
    \label{fig:dpo_training_trajectory}
    \vspace{-1em}

\end{figure*}

\section{Response sampling details}
\label{sec:response_sampling_details}

We share the full set of system prompts that we use and the process for devising them in Appendix \ref{sec:seed_system_prompts}. 
Lastly, to further diversify the set of responses and thus enhance the generalizability of the reward model, we pair these system prompts with various base \lm s, which are Falcon Instruct 7B~\cite{falcon40b}, GPT-4, and GPT-3.5. 
In addition, we also adjust the decoding temperature from 0.7, 1.0, and 1.3.  
We include the reference response from the original dataset as well, and this yields a total of 37 configurations ($4\times3\times3+1$).
By choosing two out of these 37 configurations, we have 666 possible combinations. 

\section{System prompts}
\label{sec:seed_system_prompts}

Our set of system prompts are presented in \autoref{tab:system_prompts}. 
\texttt{Simple} is the simplest prompt that simply adds ``voice" in front of assistant. 
\texttt{Medium} and \texttt{Easy} are slightly more complex variants intended for covering different levels of voice suitability between \texttt{Simple} and \texttt{Detailed}. 
\texttt{Medium} focuses on generating more conversational and compact responses while \texttt{Easy} is geared towards generating text with high readability, hence listenability, by using simple vocabulary and sentence structures. 
\texttt{Detailed} is the prompt where we place the most effort in trying to get the most consistent performance with GPT-4 to generate responses that we deemed as most suitable for speech based on the lessons from audio reporting outlined in Section \ref{sec:lessons_audio_reporting}. 

\section{Annotation Details}
\label{sec:annotation_details}

\subsection{Annotators}
\label{sec:annotator_details}

Before inviting annotators to a larger batch of annotations, each annotator was asked to complete 10 annotations that were also completed by the authors, which were manually evaluated for consistency and accuracy according to the given guidelines.
If their explanations were valid and consistent with their annotations, and the Cohen's Kappa coefficient was greater than 0.6, they were invited to a larger batch for which quality was monitored on a small sample for every 100 annotations that were completed. 
If not, we provided feedback to the annotators and asked them to do another 10, repeating this process until the annotations met our standards. 

We paid our annotators a rate that converts to an hourly wage that exceeds the minimum wage from where this study was conducted.\footnote{Details hidden for anonymity.}
We had a total of 71 unique annotators from countries with English spoken as the primary language participate in these annotations. 

Our evaluation setup shows strong interannotator agreements, as measured by Krippedorf's Alpha~\cite{krippendorff2011computing}, as shown in \autoref{tab:speech_vs_text_survey_interannotator_agreement}. Although not all of them passes the 0.8 threshold proposed by the original paper, it has been suggested that this threshold should be adjusted based on the difficulty and subjectivity of the task
~\cite{gwet2011krippendorff}. 

\begin{table}[t]
\begin{adjustbox}{width=\columnwidth,center}

    \centering
    \begin{tabular}{lrrrrrr}
    \toprule
         &  \textbf{Rel.} & \textbf{Help.} & \textbf{Correct.} & \textbf{Inform.} & \textbf{Len.} & \textbf{Under.} \\  \midrule
         \textbf{Text} & 0.988 & 0.963 & 0.978 & 0.726 & 0.600 & 0.727 \\  
         \textbf{Audio} & 0.983 & 0.910 & 0.955 & 0.785 & 0.556 & 0.765   \\ \bottomrule
    \end{tabular}
\end{adjustbox}
    \caption{Inter-annotator scores reported with Krippendorf's Alpha~\cite{krippendorff2011computing} for annotation results shared in \autoref{tab:speech_vs_text_survey_results}.}
    \label{tab:speech_vs_text_survey_interannotator_agreement}

    \vspace{-1em}

\end{table}

\subsection{Annotation interface and guidelines}
\label{sec:annotation_interface}

\begin{figure*}
    \centering
    \includegraphics[width=\textwidth]{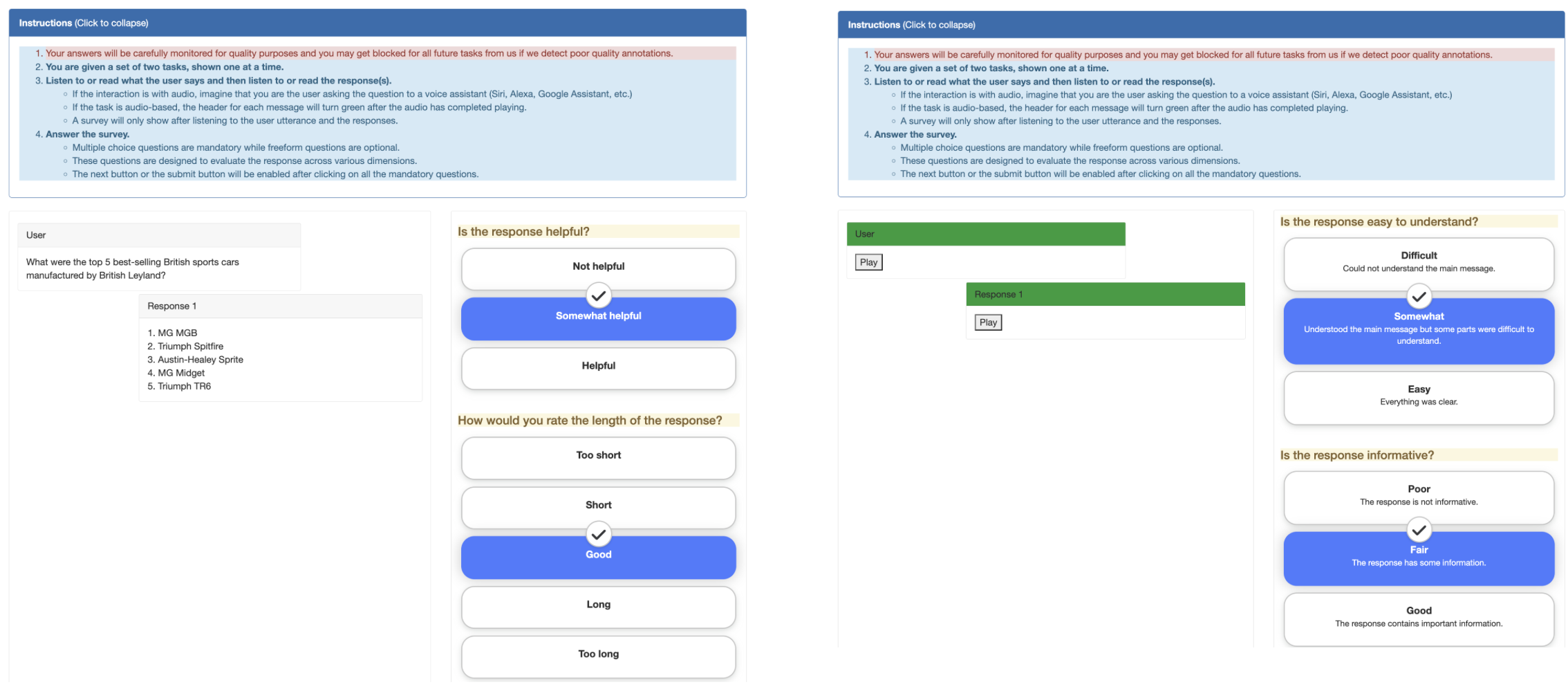}
    \caption{\textit{Left}: suitability annotation interface for text. \textit{Right}: suitablity annotation interface for audio. The same set of survey questions are asked in both cases and only the modality of the user prompt and responses are different. Note that each annotator complete a pair of both text and audio tasks so that we collect results that are better calibrated.}
    \label{fig:voice_vs_text_annotation}
\end{figure*}

The annotation interface for text vs. voice annotations is shown in \autoref{fig:voice_vs_text_annotation}. 
In order to calibrate annotation results in this comparison task, we required each annotator to complete both text-based and audio-based task, given in random order. For the audio-based task, annotators can only see the survey after they completely listen to the audio-based interaction. 

\begin{figure*}
    \centering
    \includegraphics[width=\textwidth]{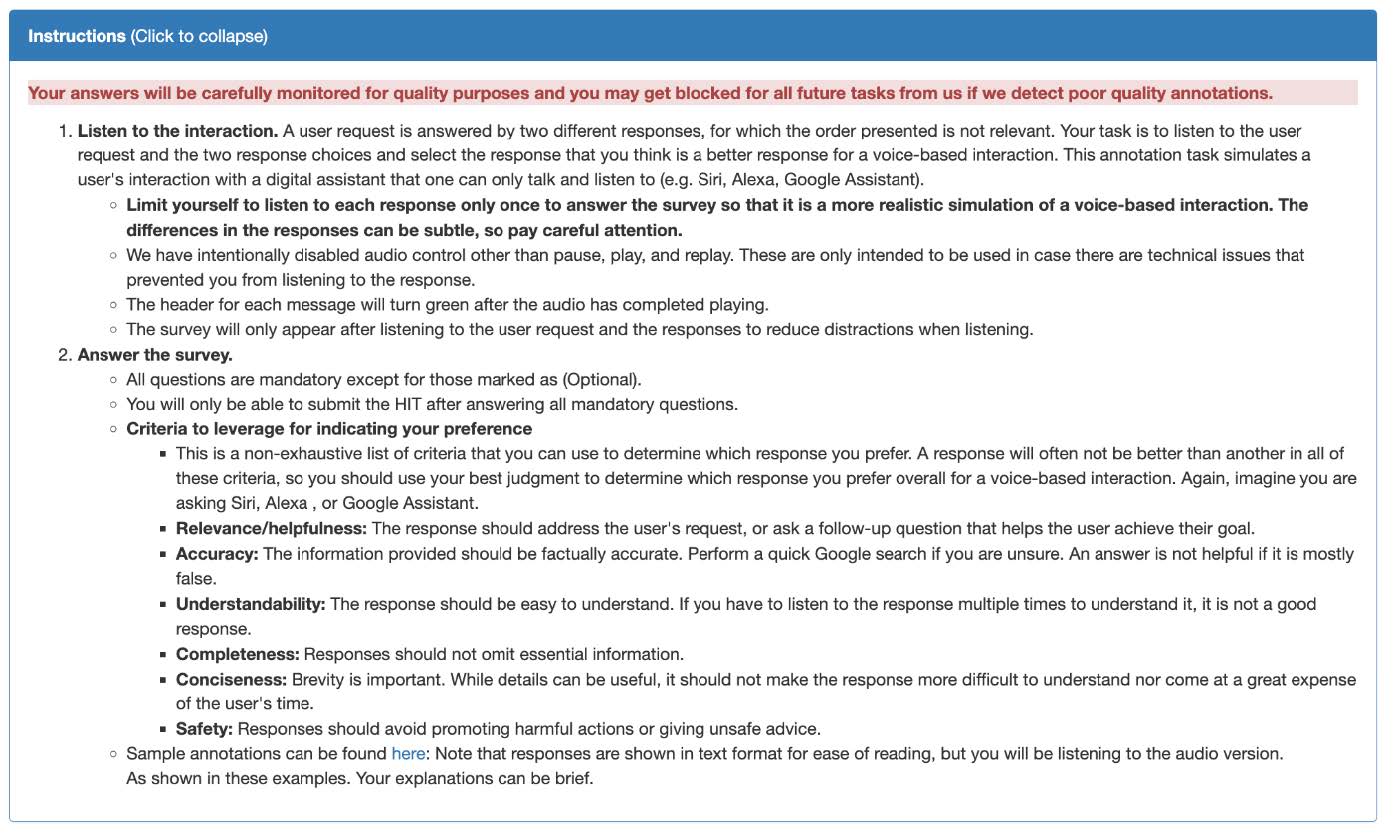}
    \caption{
        Instructions for preference annotations with responses provided as audio. 
    }
    \label{fig:audio_preference_annotation}
\end{figure*}

\begin{figure*}
    \centering
    \includegraphics[width=\textwidth]{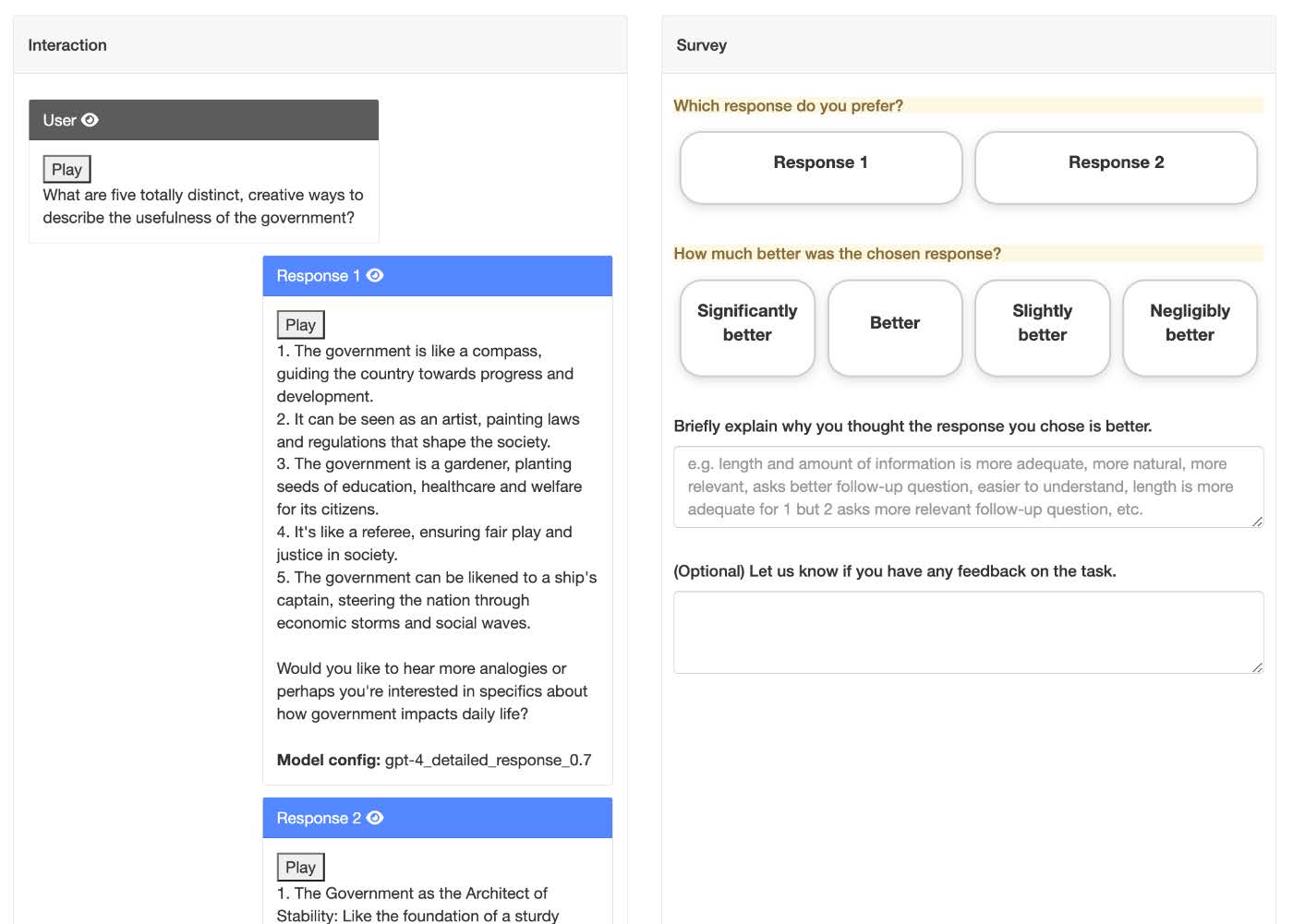}
    \caption{
        Audio preference annotation interface. The left pane contains the user prompt and two responses. The right side contains the survey that appears after the annotator listens to both responses. 
        The annotators actually do not see the text and is only available for the task distributors when monitoring completed tasks. The actual view of the annotators are the same as the right side of \autoref{fig:voice_vs_text_annotation}. 
    }
    \label{fig:preference_annotation_interface}
\end{figure*}

The interface for audio preference annotations are shown in \autoref{fig:preference_annotation_interface}. 
In order to accurately capture speech-based preferences, we mostly let the annotators and evaluators freely choose which response they considered to be more suitable for speech. 
They are only informed that they are evaluating these responses in the context of an interaction with voice assistants such as Siri, Alexa, and Google Assistant. 
Some annotators felt the task was initially too open-ended, and therefore we additionally provided a non-exhaustive criteria that they can refer to when making their decisions. 
The instructions for annotating is shown in \autoref{fig:audio_preference_annotation}.

\section{DPO vs PPO}
\label{sec:ppo_vs_dpo}

\paragraph{DPO better than PPO for OLMo, tied for Falcon (caveats)}

\begin{figure}[t]
    \centering
    \includegraphics[width=\columnwidth]{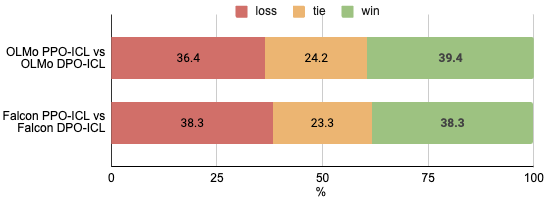}
    \caption{Head-to-head human evaluation results between best best performing PPO and DPO models for Falcon and OLMo. This figure takes the same format as \autoref{fig:human_eval_main}. 
    }
    \label{fig:ppo_vs_dpo}
\end{figure}

\autoref{fig:ppo_vs_dpo} summarizes the result of 60 pairwise comparison results, following the same methodology mentioned in \textsection \ref{sec:evaluation}.
Due to financial constraints, we limit our study to the best performing DPO and PPO results for each model, which are DPO-ICL and PPO-ICL. 
We see that DPO results are better than PPO for OLMo by a small margin and that the results are exactly tied for Falcon.

However, there are some caveats that need to be taken into account for interpreting these results.
First, there are many hyperparameters involved when training with DPO and PPO. 
Although we conducted a sweep with learning rates, batch size, DPO's beta, and various generation hyperparameters for PPO during training, there are still many other unexplored configurations that may have led to different outcomes. 

For DPO, it is straightforward to pick the best checkpoint to use for evaluation, because the checkpoint with the lowest loss closely corresponds to the checkpoint with the best accuracy during evaluation. 
However, for PPO, we found that choosing the best model based on the highest reward score or lowest loss can lead to checkpoints that produce degenerate outputs or those that converged to stubbornly abiding to a particular response pattern. For example, a checkpoint with high reward score and l. 
Therefore, we also examine sample outputs and the KL divergence from the reference model for each checkpoint, undergoing a semi-manual process to find a model that gets a high reward score while producing responses that do appear speech-suitable.

\section{Implementation and Technical Details}
\label{sec:implementation_details}

For training our reward model and finetuning Falcon Instruct 7B with DPO and PPO, we use 8 A100 GPUs.  
We use the LLaMA-Factory repository \cite{zheng2024llamafactory}, which is built on top of popular HuggingFace frameworks such as TRL and PEFT. 
We use the default training hyperparameters for both reward modeling and PPO.

\section{Qualitative Analysis}
\label{sec:qualitative_analysis}

\paragraph{Where do our finetuned models fall short?}
To understand the gap between our models and state-of-the-art \lm s, we examine the Falcon DPO-ICL vs. GPT-4-Prompt/ICL comparisons. 
We notice that even after controlling for factual knowledge, there were still instances where the responses from GPT-4-ICL were preferred because they were more relevant. 
For instance, for a user prompt that asks \textit{``What did the 0 say to the 8?''}, Falcon DPO-ICL produces a generic greeting while GPT-4-ICL understands that this is a joke and answers \textit{``The 0 said to the 8, Nice belt.''}, along with an explanation of the joke. 
Also, there are cases where \ours~ was overly succinct. 
For example, it only said \textit{``Portugal''} to \textit{``Goa used to be a colony of which nation?''}, and so GPT-4-ICL's response was preferred despite being longer as it shared related information about Goa. 
Therefore, in addition to the shortcomings in instruction-following skills due to a lack of general knowledge, including understanding of riddles and jokes, Falcon DPO-ICL's main drawback in terms from speech-suitabilty stems from the difficulty in adequately judging the right degree of brevity and amount of information.  
We share more sample comparisons and analysis in the following sections.

\subsection{Falcon DPO-ICL vs. Falcon-ICL}
\label{sec:qualitative_anaysis_vs_falcon}

We share sample comparisons between Falcon DPO-ICL and Falcon-ICL in \autoref{tab:falcon_vs_speech_falcon_qualitative_examples}. 
We show both cases where Falcon DPO-ICL was preferred and where Falcon-ICL was preferred, categorizing each with a reason, and indicating how often each reason was mentioned as part of the evaluator's explanation for their preference or was observed in the responses. 
Both models were considered more succinct, but for different instances. Falcon DPO-ICL was considered more conversational, which accounts for 28\% of the preferred cases for \ours. It also provided better follow-up questions, which occurred 37\% of the times it was the preferred choice.

\subsection{Falcon DPO-ICL vs. GPT-4-ICL}
\label{sec:qualitative_anaysis_vs_others}

We share sample comparisons between Falcon DPO-ICL and GPT-4-ICL in \autoref{tab:speech_falcon_vs_gpt_4_prompt}. 
The most noticeable reason for GPT-4-ICL's dominance over Falcon DPO-ICL was in that it was better at balancing length and actually addressing the user's request, while Falcon DPO-ICL was good at generating natural sounding text without non-vocalizable content but sometimes would not answer the request and go directly to a follow up question.

\subsection{Is prompt engineering with larger models sufficient for speech-suitable responses?}

The dominant performance of GPT-4-ICL shown in \autoref{fig:gpt_results} raise the question whether larger \lm s with only prompt engineering is sufficient for reliably producing speech-suitable responses. 
There are several practical arguments for why this approach is not sufficient. 

\textit{(i) Inconsistency in following instructions:} even GPT-4-ICL, the model with the best win rate over \textbf{Falcon DPO-ICL}, generated lists or non-vocalizable content for 13.6\% of its responses, despite being explicitly told not to do so. On the other hand, \textbf{Falcon DPO-ICL} had only 1.2\% such responses. 
\textit{(ii) Speech suitability is context-dependent:} as illustrated in Section \ref{sec:human_evaluation_results} and \ref{sec:automatic_eval_results}, speech-suitability is context-dependent and not always about being more simple or concise. 
A naive solution would be to expand the system prompt to catch as many edge cases as possible, but recent work has found that detailed system prompts can lead to worse reasoning skills, which could lead to worse performance in unexpected use cases~\cite{gupta2023bias}. 
In conclusion, these practical issues endorse speech-centric finetuning approaches such as \ours~ to develop \lm s that can more reliably generate responses that capture the nuances of speech preferences without relying on extensive prompt engineering.

\section{Use of AI assistants}

Github Copilot was used throughout the coding process for this work. None was used to assist with research or writing. 

\begin{table*}[t]
\small
\begin{adjustbox}{width=\textwidth,center}

\centering

    \begin{tabular}{p{0.15\textwidth}p{0.17\textwidth}p{0.03\textwidth}p{3cm}p{5cm}p{5cm}}
    \toprule
    \textbf{Preferred} & \textbf{Category} & \textbf{\%} & \textbf{User prompt} & \textbf{Falcon DPO-ICL response} & \textbf{Falcon-ICL response} \\
    \midrule  

    \multirow{25}{*}{\textbf{Falcon DPO-ICL}}& More conversational & 28\% & Do RVs have bathrooms? & Yes, RVs typically have bathrooms. Would you like more information about it? & Generally, RVs do have bathrooms. \\ \cmidrule(lr){2-6}

    & More succinct & 26\% & Why Men love whisky & Men love whisky because it is a drink that is full of flavour and it can be enjoyed both on its own and with food. Would you like to know more about it? & Whisky is a distilled spirit made from fermented grain mash. It is typically matured in oak barrels, giving it a unique smoky flavor. The drink is known for its versatility and complexity. There are many reasons why a person may choose to enjoy a glass of whisky, but it is likely due to the taste, the relaxing effects, the sense of sophistication associated with the drink, and its ability to pair well with food. Whisky has a long history dating back to ancient times, and it has been enjoyed by many cultures across the globe. It is often thought to be a masculine drink due to its strong and sometimes complex taste. Do you want any more information on this topic? \\ \cmidrule(lr){2-6}
    
    & Better follow-up & 37\% &  Who maintains the Tottori Expressway? & The Tottori Expressway is maintained by the Tottori railway company and the Tottori prefectural government. Would you like more information on this topic?& The expressway is maintained by the Tottori Prefecture and the Tottori Railway Company. \\ \cmidrule(lr){2-6}
    
    & No list & 12\% & Think up of three different animals that lay eggs &Three animals that lay eggs are chickens, ducks, and turkeys. & 1. Chicken 2. Frog 3. Turtle\\
    
    \midrule 
    \multirow{25}{*}{\textbf{Falcon-ICL}}
    & More comprehensive & 19\% & When is the best age for kids to approach sailing? & Generally, it is considered a good age for kids to start learning about sailing between the ages of 6 and 8 years old. & In my opinion, kids as young as 5 can approach sailing with appropriate guidance from adults. The child should learn the basics of sailing such as wind direction, how to maneuver the boat, and safety practices. It is recommended to start with a small boat that is more stable for beginners. As the child grows and improves, they can move to larger boats and learn the more advanced skills. The age limit for learning to sail would depend on the individual child's maturity level and the parent's comfort level. It is recommended for the child to have some basic knowledge of water safety and understanding the basic rules of the waterway before they are allowed to go out on the water. \\ \cmidrule(lr){2-6}
    & More succinct & 16\% & What does a lawyer do? & A lawyer is a professional who advises and represents individuals, businesses, and organizations on legal issues and disputes. Lawyers are generally responsible for providing legal support and guidance, researching and interpreting applicable laws, and drafting legal documents such as contracts and pleadings. & A lawyer advises clients on legal matters and represents them in court proceedings. They help ensure that the law is upheld and advocate for their clients' rights. \\ 

     \bottomrule
    \end{tabular}
    \end{adjustbox}
    \caption{
        Representative sample comparisons for Falcon DPO-ICL vs. Falcon-ICL. Category is an nonexhaustive and nonexclusive reason for why the winning response was preferred, either given as an explanation by the evalutors or was observed during qualitative analysis. \% indicates how often each category was observed in the preferred cases. 
    }
    \label{tab:falcon_vs_speech_falcon_qualitative_examples}
\end{table*}

\begin{table*}[t]
\small
\centering
\resizebox{\textwidth}{!}{
\renewcommand{\arraystretch}{1}
    \begin{tabular}{p{0.15\textwidth}p{0.2\textwidth}p{0.03\textwidth}p{3cm}p{5cm}p{5cm}}
    \toprule
    \textbf{Preferred} & \textbf{Category} & \textbf{\%} & \textbf{User prompt} & \textbf{Falcon DPO-ICL response} & \textbf{GPT-4-ICL response} \\
    \midrule  

    \multirow{25}{*}{\textbf{Falcon DPO-ICL}}& No list & 29\% & What to do in Seattle in 48 hours? & There are many exciting activities in Seattle that you can do in 48 hours. Some popular ones include taking a scenic boat tour, hiking in the nearby parks, and exploring the unique neighborhoods like Ballard and Wallingford. Would you like more details about these activities? & In 48 hours, there are several things you can do in Seattle:

1. Visit the Pike Place Market, which is a bustling farmers market famous for its fresh produce and unique crafts.

2. Check out the stunning views from the top of the Space Needle, a Seattle landmark.

3. Explore the Chihuly Garden and Glass, where you'll see spectacular glass sculptures.

    \\ \cmidrule(lr){2-6}
    & More natural & 33\% & How can I initiate conversation with a stranger? &   It can be challenging to initiate conversation with a stranger, but there are a few ways to make it easier. One way is by asking a question about something they are doing or wearing. Another way is by complimenting something they are doing or have done, like a hobby or a piece of clothing. Would you like more tips? & Step one, approach the person kindly with a smile. Second, make a comment about something in your immediate environment. This could be about the weather, a piece of art, or even a sports team logo. Third, introduce yourself and ask about their interests or hobbies. Are you ready for further advice, or would you prefer to practice these initial steps first?\\
    \\
    \midrule 
    \multirow{28}{*}{\textbf{GPT-4-ICL}} & Knowledgeable / More relevant & 44\% & What did the 0 say to the 8? & Hey there! I'm just a little 0.& The 0 said to the 8, 'Nice belt.' That joke is referring to how the number 8 looks like a 0 with a belt tightened around its middle. Do you want to hear another fun joke about numbers? \\ \cmidrule(lr){2-6}
    & More comprehensive & 52\% &Is Spain a good place to live? & Spain is a wonderful place to live. Would you like to know more about it? & Spain can be a great place to live, depending on your personal preferences. It offers a warm climate, rich history and culture, wonderful cuisine, and in many areas, a relatively low cost of living. It also has universal healthcare. However, Spain does have high unemployment, and knowing Spanish is important for integrating and finding a job. As each person's criteria for a 'good' place to live can be different, what are you looking for in a place to live?\\ \cmidrule(lr){2-6}
    & Provides answer & 32\% &Can you help me avoid credit card debt? &Sure, I can offer you some tips on avoiding credit card debt. Would you like to know more about them?  & Absolutely, here are three simple strategies to avoid credit card debt:

1. Pay your bill on time each month to avoid interest and late fees.

2. Try to pay more than the minimum due when you can. This will help reduce your balance faster.

3. Use your credit card only for purchases you can afford to pay off each month.

Would you like more advice on this? \\

     \bottomrule
    \end{tabular}}
    \caption{
        Representative sample comparisons for Falcon DPO-ICL vs. GPT-4-ICL. The schema of this table is the same as \autoref{tab:falcon_vs_speech_falcon_qualitative_examples}. While GPT-4-ICL tends to generate longer respones, it is at the expense of providing the amount of information that the evaluators often found necessary for a comprehensive understanding. 
    }
    \label{tab:speech_falcon_vs_gpt_4_prompt}
\end{table*}

\end{document}